\pdfoutput=1

\documentclass[11pt]{article}

\usepackage[]{EMNLP2023}

\usepackage{times}
\usepackage{latexsym}
\usepackage{fancyhdr}

\usepackage[T1]{fontenc}

\usepackage[utf8]{inputenc}

\usepackage{microtype}

\usepackage{inconsolata}

\usepackage{graphicx} 

\usepackage{url}  

\title{Text2Topic: Multi-Label Text Classification System for Efficient Topic Detection in User Generated Content with Zero-Shot Capabilities}

 \author{
    {\bf Fengjun Wang$^{\dagger}$,}
    {\bf Moran Beladev$^{\dagger}$,}
    {\bf Ofri Kleinfeld,}
    {\bf Elina Frayerman,}
    \\
    {\bf Tal Shachar,}
    {\bf Eran Fainman,}
    {\bf Karen Lastmann Assaraf,}
    {\bf Sarai Mizrachi,}
    {\bf Benjamin Wang} 
\\
Booking.com\\
\{fengjun.wang, moran.beladev, ofri.kleinfeld, elina.frayerman, tal.shachar,\\
eran.fainman, karen.lastmannassaraf, sarai.mizrachi, benjamin.wang
\}@booking.com \\
    $^{\dagger}$These authors contributed equally to this work}

\begin{document}
\thispagestyle{fancy}
\chead{This paper was accepted to EMNLP 2023. Please reference it instead once published}
\rhead{}
\maketitle
\begin{abstract}

Multi-label text classification is a critical task in the industry. It helps to extract structured information from large amount of textual data. 
We propose Text to Topic (Text2Topic), which achieves high multi-label classification performance by employing a Bi-Encoder Transformer architecture that utilizes concatenation, subtraction, and multiplication of embeddings on both text and topic. 
Text2Topic also supports zero-shot predictions, produces domain-specific text embeddings, and enables production-scale batch-inference with high throughput. 
The final model achieves accurate and comprehensive results compared to state-of-the-art baselines, including large language models (LLMs).

In this study, a total of 239 topics are defined, and around 1.6 million text-topic pairs annotations (in which 200K are positive)
are collected on approximately 120K texts from 3 main data sources on Booking.com.
The data is collected with optimized smart sampling and partial labeling. The final Text2Topic model is deployed on a real-world stream processing platform, and it outperforms other models with $92.9\%$ micro mAP, as well as a $75.8\%$ macro mAP score. We summarize the modeling choices which are extensively tested through ablation studies, and share detailed in-production decision-making steps. 
\end{abstract}




\section{Introduction}
In the digital age, large-scale  online travel platforms (OTPs) face the challenge of effectively extracting valuable insights from massive volumes of textual data. 
Such an OTP can get hundreds of millions of customer reviews in one year, so structured insights are crucial for comprehending customer behavior and making data-driven decisions in order to improve the overall travel experience. 
One application example is to find the top facilities for each hotel, by extracting information from the positive reviews, which can lead to better accommodation recommendations. Similarly, understanding travel destination themes, such as romantic getaways, city trips, or family trips can enhance destination recommendations.  
In this study, we research use cases from Booking.com and define in total 239 valuable topics. Each topic is set with a topic name and a topic description, to better match the natural customer language and for optimal model training results. The main data source is user-generated content on Booking.com, including customer reviews and forum posts from hotel owners and travelers. 

Developing an architecture that ensures high accuracy, scalability for a large number of topics, low cost and low latency on real-world inference is of utmost importance. 
Sentence-BERT \cite{sbert} extends BERT \cite{bert} for sentence-level embeddings, achieving impressive performance on tasks like sentence similarity and semantic retrieval. Multilingual Universal Sentence Encoder for Semantic Retrieval (MUSE) \cite{muse}, a multilingual extension of the Universal Sentence Encoder \cite{use}, enables cross-lingual semantic retrieval and provides multiple open-source models. 
Though there are also other state-of-the-art approaches, the two methods above are prevalent in real industry applications, due to the computational efficiency, high and robust in-domain performance by fine-tuning, zero-shot ability, and strengths in scalability.  

Our proposed Text2Topic framework adopts a fine-tuning approach upon pre-trained language models. Specifically, we employ the bi-encoder transformer \cite{vaswani2017attention} architecture proposed by Sentence-BERT, which allows separate injection of the text and topic information, as Section \ref{sec:arch} shows. 
This architecture not only enables the model to have zero-shot capabilities (handle new topics for inference) but also exhibits text embedding abilities. 
By leveraging the strengths of the pre-trained language model and incorporating topic-specific information, the model effectively addresses the challenge of topic detection. 
The paper's contributions are summarized as follows:  
\begin{itemize}
    \item We propose a practical Text2Topic framework for the efficient extraction of topics from texts.  
    \item We share the model development core findings, including multiple model architectures' comparison, training one universal model versus dedicated models per data source, outperforming against baselines (MUSE, GPT-3.5). 
    \item We share efficient and practical dataset annotation strategies with smart model-based sampling, as described in Section \ref{sec:in-house-dataset}. 
    \item We provide zero-shot capability for unseen classes, which performs better than MUSE when the unseen class is in the travel domain. 
    \item We detail the real-world use cases in Section \ref{sec:use_case}, and deployment decisions in Section \ref{sec:deploy}. 
\end{itemize}




\section{Architectures}\label{sec:arch}



\begin{figure*}[ht]
 \centering
 \includegraphics[width=1\textwidth]{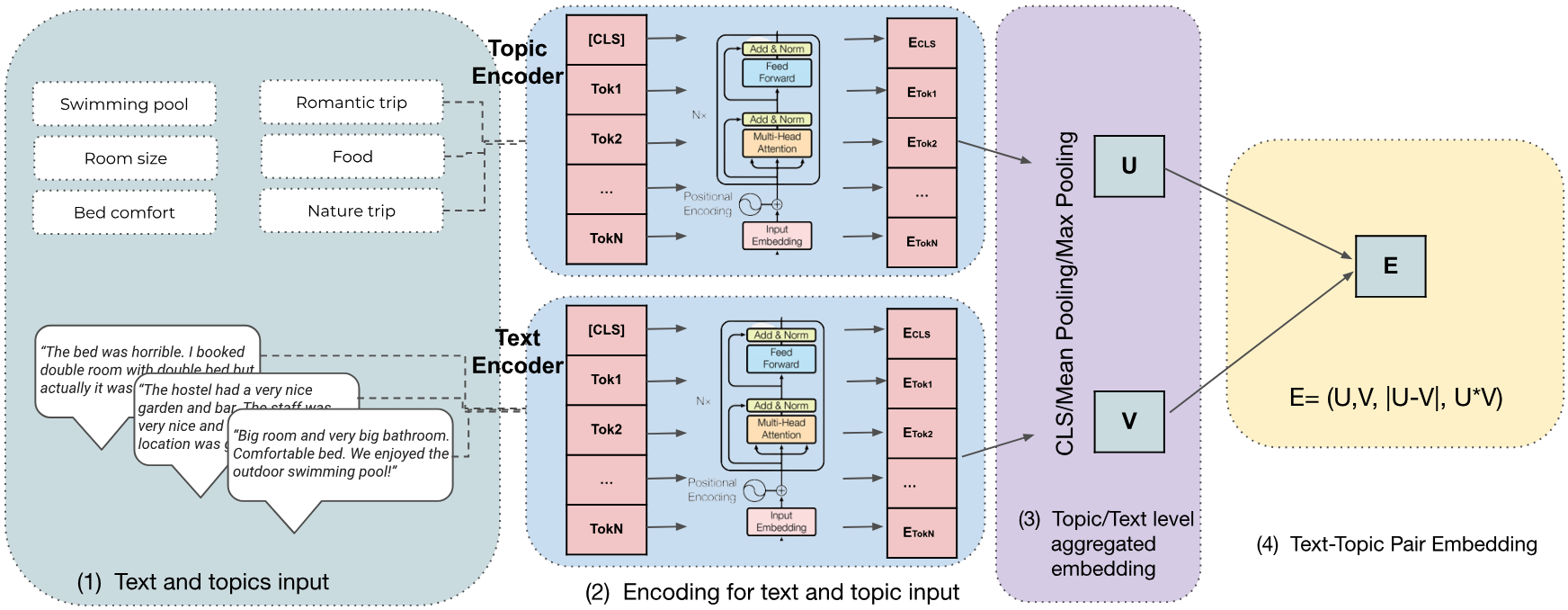}
    \caption{Bi-Encoder Concatenation. (1) Input Text-Topic pairs. (2) Encode each text and topic with transformer encoder (the two encoders share weights). (3) Aggregate each of the two text-topic token embeddings into one single vector to represent the topic ($U$) or the text ($V$) using CLS/mean-pooling/max-pooling. (4) Combine the two embeddings into one representation of the pair relationship, $E$. Then feed $E$ into two feedforward layers to get logits output, where BCE loss is applied on. 
    For inference, we broadcast topics embeddings to pair with text embeddings and score each pair.
    }
    \label{fig:model_architecture}
\end{figure*}

In this study, we research 3 main architectures. For each text-topic (with topic description), we know one binary ground truth for this pair, and perform: 

\begin{itemize}
    \item Cross-encoder: the text and topic description are tokenized and concatenated as one input with \verb\[SEP\] separator 
    (``\verb"TEXT[SEP]TOPIC"''), 
    then passed into the transformer encoder and a classification head to derive logits, where Binary Cross Entropy (BCE) loss is applied.  

    \item Bi-encoder Concatenation (Figure \ref{fig:model_architecture}): we first generate a pair of embeddings $(U, V)$ both as dimension $d_{model}$, where $U$ is the topic description embedding 
     and $V$ is the text embedding. Then we feed $E$ (the embedding concatenation, subtraction and multiplication) into 2 feedforward layers (${FFN_1} \in R^{d_E \times d_{model}}$, ReLU activation, and dropout, and then ${FFN_2} \in R^{d_{model} \times 1}$), and finally apply BCE loss.
    \item Bi-encoder Cosine: similar to the Figure \ref{fig:model_architecture}, but at step 4, instead of embedding combination, we apply cosine similarity directly on $U$ and $V$, and then apply mean-squared-error loss as the objective function. 
    

\end{itemize}


The bi-encoder architecture has 3 benefits over cross-encoder: 1) Low inference time-complexity: we pre-calculate and cache all topic embeddings, only embed each text once and repeat the text vector to score on all topics. Given $N$ as the number of texts and $T$ as the number of topics, to get $N \times T$ predictions, the bi-encoder needs $O(N + T)$ encoding operations, while it is $O(N \cdot T)$ for the cross-encoder. 2) In-house embedding: bi-encoder enables us to have the text part embeddings, 
that can be used as features for other tasks. 3) For the same base model, the bi-encoder allows longer text input since text and topic are embedded separately. 

All the above architectures can easily extend to include new topics for training, and also have \textbf{zero-shot} possibility when the unseen topic is well-defined with a description. For all architectures, we experiment with three pooling strategies \cite{sbert}: using the output of the \verb\[CLS\]; computing the mean or max of output vectors on all tokens (mean-pooling, max-pooling).

\section{In-house Dataset Construction}  \label{sec:in-house-dataset}
With hundreds of topics, the annotation becomes challenging: how to define, merge, and distinguish topics; how to decide annotation volume and candidate texts per topic and reduce cost. This section shows how we tackle them by smart model-based sampling and partial labeling. 

\subsection{Annotation Volume Estimation} \label{sec:anno_number}
We start with a proof-of-concept stage, where 43 topics are pre-defined and annotated by domain experts on 12K texts. With the data, we run multiple cross-encoder model training by increasing the number of positive annotations per topic in the training data. Figure \ref{fig:num_anno} in the Appendix shows that for most topics, the mAP metrics saturate at 200 positive annotations, which reflects basic guidance. 


\subsection{Topic Definitions}
In the end, we define 239 topics from user researches, covering broader topics such as trip types (romantic trip, city trip etc.), travel activities (surfing, hiking etc.), and specific user needs such as hotel facilities (garden, balcony etc.). Each topic has a name and a description which is refined with the help from LIME (see Section \ref{sec:lime}). 

\subsection{Smart Sampling and Partial Labeling} \label{sec:active_learning}
In our corpus, text is typically short and contains low number of topics, so we apply partial labeling instead of full annotation. 
The 239 topics are split into 38 multi-choice question groups (e.g., food topics are in one group).  
With the best 43-topic model (from the training as Section \ref{sec:anno_number} shows), we apply it to predict on a large corpus, detecting 239 topics and generating text-topic scores. Notably, the prediction results for the unseen 196 topics are generated by zero-shot. 


With the predictions, we perform smart text sampling: 
1) Firstly, for each topic, we do probability-weighted sampling on the texts whose scores pass a threshold,  
and assign selected texts to the multi-choice group which contains that topic.
Figure \ref{fig:annotation_example} in the Appendix shows an example for a text that passes the threshold for the "romantic trip" topic and was assigned with the group of topics that contains the "romantic trip" topic. 
2) In addition, to avoid annotation bias, each selected text is also assigned to one random group (besides the already assigned relevant groups). For example, the text in Figure \ref{fig:annotation_example} is also assigned to another group randomly. 
3) Besides the model-based text sampling, we also randomly sample some texts from corpus and randomly assign them to groups. 

We use AWS SageMaker Ground Truth as the platform for the annotations collection, and leverage on some of the MuMIC \cite{wang2023mumic} annotation pipelines and strategies (majority voting etc.).  
We recruit specialized annotators to form one auditing team, and multiple worker teams. After a knowledge transfer phase, we start the production phase where each task is done by 3 workers, and the labels are inferred by majority voting. 
The auditing team performs regular performance checks
and we get > 95\% annotation accuracy. 
Finally, we collected almost 1.6 million annotations at a low cost, with 200K of them being positive (which is 12.5\%). These annotations are gathered from approximately 120K unique multilingual texts, including English, German, French, Russian, etc., from guests, travelers, and property owners, sourced from reviews, the travel community, and partner hub.




\section{Experimental Setup}
All experiments are performed on a computation instance equipped with 1 NVIDIA Tesla T4 Tensor Core GPU, 4 vCPU, and 16GB RAM. 
The experiments have the following general settings: bert-base-multilingual-cased model as the pre-trained base model; fine-tuning all layers; mixed-precision training; batch size of 12 text-topic pairs, with gradient accumulation steps as 8; weight decay (on all weights that are not gains or biases) with coefficient 0.01; AdamW optimizer \cite{weight_decay_adam}; initial learning rate 1e-5 with a linear scheduler; allowing maximum 6 epochs with early stopping patience as 3 steps, and warm-up steps as 10K. 
We apply stratified sampling (on topic frequency) on the texts, and get training/validation/test sets, with a ratio of 70/15/15 respectively. 



\subsection{Evaluation Metrics}

Given $T$ topics, and $N$ texts, the ground truth and the predictions can both be represented as a matrix with size $N \times T$ \footnote{With partial labeling, the matrix has null values, and we filter them out accordingly when do metrics calculation.} . 
We use the below metrics as the main evaluation criterion \cite{eval_paper}:

\begin{itemize}
    \item 
    Average Precision (AP) per class:
\begin{equation}
    AP_j =  \sum_{i=1}^{N} p_j(i)\Delta r_j(i) 
\end{equation}

where $p_j$ is the precision of class $j$, and $r_j$ is the recall of class $j$.  
It is equivalent to the area under the precision-recall curve per class. 

\item 
macro Mean Average Precision
\begin{equation}
    macro \: mAP = \frac{1}{T} \sum_{j=1}^{T} AP_j
\end{equation}

which is the unweighted average of AP on all classes, treating each class equally. 

\item 
weighted Mean Average Precision
\begin{equation}
    weighted \: mAP = \frac{1}{\sum_{j=1}^{T} {NP_j}} \sum_{j=1}^{T} AP_j \cdot NP_j
\end{equation}

where $NP_j$ is the number of positive samples of class $j$.

\item
micro Average Precision (global-based)
\begin{equation}
    micro \: mAP = \sum_{i=1}^{N \cdot T} p(i)\Delta r(i) 
\end{equation}


\end{itemize}

\subsection{Baselines}
\textbf{MUSE} \cite{muse}: Google provides multiple versions of MUSE models, and we use the ``multilingual-large-3'' one \footnote{https://tfhub.dev/google/universal-sentence-encoder-multilingual-large/3}. The model covers the languages we have in the dataset, is trained with multi-task learning on Transformer architecture, 
and is optimized for multi-word length text. 
Given a text, MUSE generates a 512-dimensional vector as the embedding. For each text-topic pair, we calculate the cosine similarity on text embedding and topic embedding as the model prediction score. 

\noindent \textbf{GPT-3.5}: we choose the gpt-3.5-turbo-0301, which supports a maximum 4K token context length. 

\section{Experimentation Findings} \label{sec:exp_finding_sec}

\subsection{Final Model Performance}


\begin{table}[]
\centering
\resizebox{1\columnwidth}{!}{%
\begin{tabular}{lccc}
\hline
                                                                       \textbf{method}       & \textbf{\begin{tabular}[c]{@{}c@{}}micro \\ mAP\end{tabular}} & \textbf{\begin{tabular}[c]{@{}c@{}}macro\\ mAP\end{tabular}} & \textbf{\begin{tabular}[c]{@{}c@{}}weighted\\ mAP\end{tabular}} \\ \hline
MUSE                                                                               & 37.9                                                         & 41.2                                                        & 60.4                                                           \\ \hline
Cross-encoder                                                                    & \textbf{94.7}                                                         & \textbf{81.4}                                                        & \textbf{92.8}                                                           \\ \hline
\begin{tabular}[c]{@{}l@{}}Bi-encoder (cosine)\end{tabular}                      & 89.4                                                         & 71.4                                                        & 88.5                                                           \\ \hline
\begin{tabular}[c]{@{}l@{}}Bi-encoder (concat)\end{tabular}                      & 84.3                                                         & 62.5                                                        & 80.3                                                           \\ \hline
\begin{tabular}[c]{@{}l@{}}Bi-encoder (concat, sub)\end{tabular}              & 91.4                                                         & 72.6                                                        & 89.5                                                           \\ \hline
\begin{tabular}[c]{@{}l@{}}Bi-encoder (concat, sub, mult)\end{tabular} & \textbf{92.9}                                                         & \textbf{75.8}                                                        & \textbf{91.0}                                                           \\ \hline
\end{tabular}
}
\caption{Performance comparison of Text2Topic, on the test set. All metrics are in \%.}
\label{tab:perf_arch}
\end{table}

Table \ref{tab:perf_arch} compares performance across multiple model architectures and MUSE baseline. We perform hyperparameter tuning on all methods, report the highest reachable performance and find all of them beat the MUSE baseline
\footnote{In Text2Topic hyperparamter tuning, we find the bi-encoder cosine architecture prefers mean-pooling, while bi-encoder concat models prefer [CLS] embedding. }. 
The cross-encoder outperforms all other architectures because it learns the topic-text relation attention layer by layer inside the transformer. Generally, the bi-encoder concatenation method is better than simple cosine similarity architecture, and the embedding subtraction and multiplication are both necessary.  

It's worth mentioning that for all methods, the model training typically can saturate at around 2nd or 3rd epoch, which takes less than one day for one model training. The cross-encoder has the highest performance but with too high inference time complexity, so we choose the ``bi-encoder (concat, sub, mult)'' one for production and refer it as ``bi-encoder concat'' model in this paper. 

\subsection{Train One Model or Three Models?}
As mentioned in Section \ref{sec:active_learning}, there are 3 data sources. In the dataset, customer reviews occupy more than 70\% data, and have almost all 239 topics, while the other 2 sources both have less than 100 topics. 
Should we train one model on all data or 3 models per dataset?  
Table \ref{tab:cross-dataset} shows the ablation results - under the same modeling set-up, we train and evaluate models on each source. Training on all datasets yields the best performance, we expect the model to learn patterns from all sources and gain better generalization ability. This decision also makes the model management easier.




\begin{table*}[]
\centering
\resizebox{1\textwidth}{!}{%
\begin{tabular}{l|ccc|ccc|ccc|ccc}
\hline
    & \multicolumn{3}{c|}{train on all data (all)}                                                                                                                          & \multicolumn{3}{c|}{train on customer review (rev)}                                                                                                                   & \multicolumn{3}{c|}{train on partner hub (ph)}                                                                                                                        & \multicolumn{3}{c}{train on travel community (tc)}                                                                                                                    \\ \hline
    & \begin{tabular}[c]{@{}c@{}}macro \\ mAP\end{tabular} & \begin{tabular}[c]{@{}c@{}}micro \\ mAP\end{tabular} & \begin{tabular}[c]{@{}c@{}}weighted \\ mAP\end{tabular} & \begin{tabular}[c]{@{}c@{}}macro \\ mAP\end{tabular} & \begin{tabular}[c]{@{}c@{}}micro \\ mAP\end{tabular} & \begin{tabular}[c]{@{}c@{}}weighted \\ mAP\end{tabular} & \begin{tabular}[c]{@{}c@{}}macro \\ mAP\end{tabular} & \begin{tabular}[c]{@{}c@{}}micro \\ mAP\end{tabular} & \begin{tabular}[c]{@{}c@{}}weighted \\ mAP\end{tabular} & \begin{tabular}[c]{@{}c@{}}macro \\ mAP\end{tabular} & \begin{tabular}[c]{@{}c@{}}micro \\ mAP\end{tabular} & \begin{tabular}[c]{@{}c@{}}weighted \\ mAP\end{tabular} \\ \hline
all & \textbf{71.4}                                        & \textbf{89.4}                                        & \textbf{88.5}                                           & 66.1                                                 & 87.0                                                 & 86.6                                                    & 33.5                                                 & 23.1                                                 & 53.4                                                    & 32.6                                                 & 34.9                                                 & 55.4                                                    \\ \hline
rev & \textbf{71.7}                                        & \textbf{90.3}                                        & \textbf{89.4}                                           & 70.8                                                 & 90.1                                                 & \textbf{89.4}                                           & 32.3                                                 & 22.5                                                 & 54.7                                                    & 31.3                                                 & 34                                                   & 55.3                                                    \\ \hline
ph  & \textbf{65.2}                                        & 73.2                                                 & 72.8                                                    & 28.2                                                 & 17.5                                                 & 35.1                                                    & 58.7                                                 & \textbf{73.7}                                        & \textbf{72.9}                                           & 27.4                                                 & 21.1                                                 & 39.2                                                    \\ \hline
tc  & \textbf{57.9}                                        & 68.7                                                 & 70.4                                                    & 40.9                                                 & 34.4                                                 & 49.1                                                    & 27.9                                                 & 20.6                                                 & 41.3                                                    & 56.4                                                 & \textbf{72.7}                                        & \textbf{72.0}                                           \\ \hline
\end{tabular}
}
\caption{Cross-dataset model training experimentation results, on test set, with bi-encoder cosine. We mark the highest scores as bold for each evaluation metric.}
\label{tab:cross-dataset}
\end{table*}

\subsection{Zero-Shot Evaluation}
We randomly split all topics into 5 groups, and then each time train 4 groups and evaluate the zero-shot ability on the remaining one group. 
Table \ref{tab:zero-shot-all} provides an overall performance comparison, and we see bi-encoder concat model performs the best. 
In the Appendix, Figure \ref{fig:zero_shot} and Figure \ref{fig:zero_shot_rank_on_muse} depict topic-level performance, where the bi-encoder concat has the best zero-shot ability in most topics, and Table \ref{tab:zero-shot-top-50} in the Appendix shows the aggregated performance on popular topics. 
In general, we can say that the Text2Topic keeps a balance between learning new capabilities and exposing existing capabilities. 

\begin{table}
\resizebox{1\columnwidth}{!}{
\begin{tabular}{lccc}
\hline
\textbf{method}                & \textbf{\begin{tabular}[c]{@{}c@{}}macro \\ mAP\end{tabular}} & \textbf{\begin{tabular}[c]{@{}c@{}}weighted \\ mAP\end{tabular}} & \textbf{\begin{tabular}[c]{@{}c@{}}macro \\ F1 score\end{tabular}} \\ \hline
MUSE                           & 41.2                                                          & 60.4                                                             & 46.4                                                                 \\ \hline
Bi-encoder (cosine)            & 35.8                                                          & 64.1                                                             & 41.3                                                                 \\ \hline
Bi-encoder (concat, sub, mult) & \textbf{46.8}                                                 & \textbf{71.1}                                                    & \textbf{51.1}                                                        \\ \hline
\end{tabular}
}
\caption{Zero-shot overall test-set performance on all topics. We search the
best F1 across all thresholds per topic, and then get macro averaged F1 across topics. 
}
\label{tab:zero-shot-all}
\end{table}


\subsection{Comparison with GPT-3.5} \label{sec:gpt_exp}
Considering the GPT-3.5 context length, we select 24 topics for the evaluation, covering 3 representative groups: food, trip types, and room conditions. 
With multiple prompt iterations, we find the few-shot prompting is necessary because it can regulate output format by showing examples. 
We finally get two best prompts: 8-shot and 38-shot. Both prompts include the 24 topic definition list with descriptions, and Chain-of-Thought (CoT) \cite{wei2023chainofthought} rules: ask the model to quote each part of the text, infer topics, and then output a topic list. 
Besides topic description and CoT rules, the 8-shot prompt (around 1700 tokens) has 3 text examples, covering 8 positive annotations on 8 topics; while the 38-shot (around 2900 tokens) has 16 examples, covering 38 positive annotations on 24 topics. 

\begin{figure*} 
    \centering
    \includegraphics[width=0.9\textwidth]{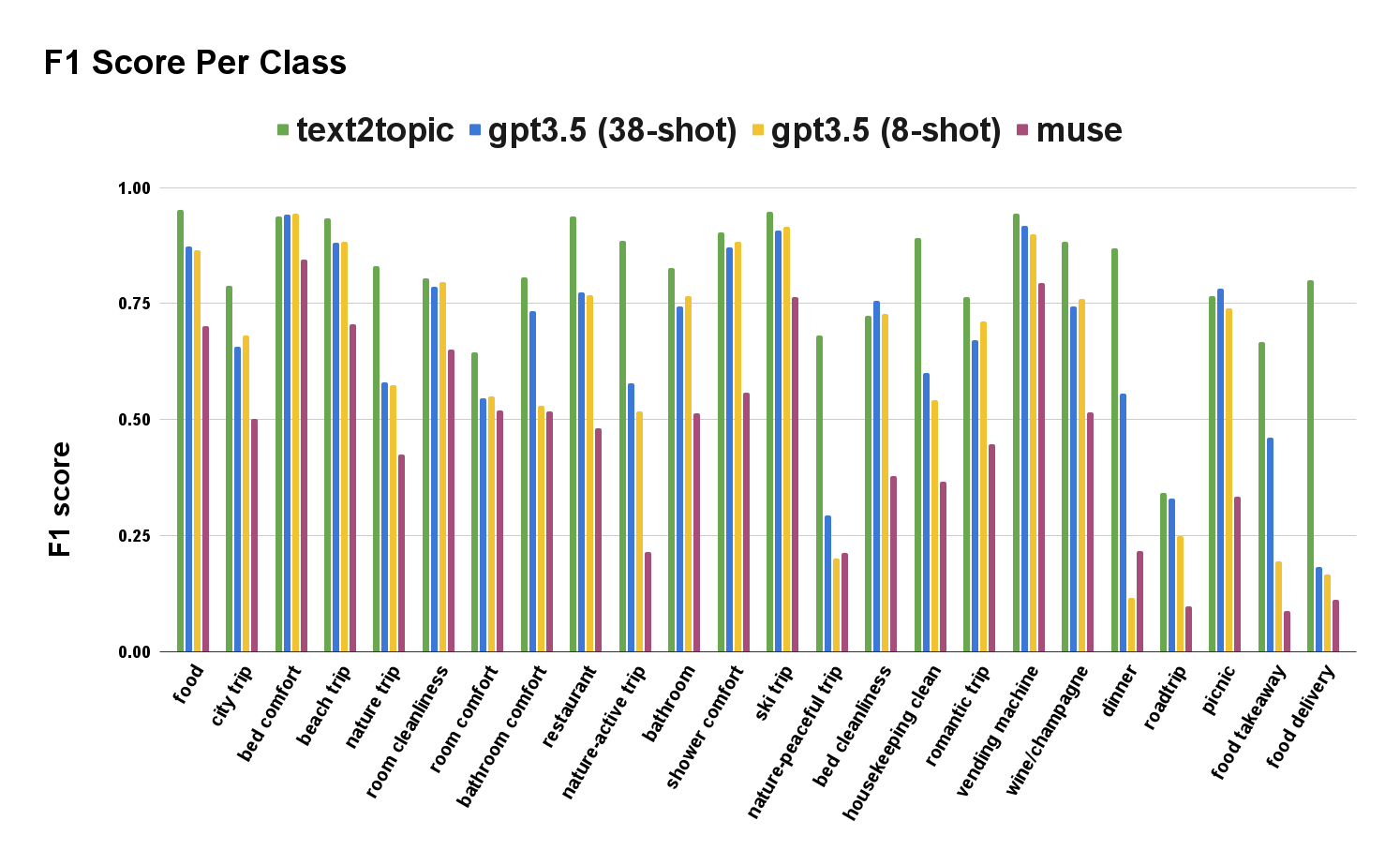}
        \caption{F1 score on each topic, on the test set. The x-axis represents the 24 topics, ordered by topic popularity (support-1). For Text2Topic and MUSE, we search the best F1 score across all thresholds for each topic. }
    \label{fig:llm_f1}
\end{figure*}

Figure \ref{fig:llm_f1} shows that Text2Topic (our bi-encoder concat model) performs the best in almost all topics, and the 38-shot prompt slightly outperforms the 8-shot. This indicates that when already having clear topic descriptions in the prompt, adding more examples is not always powerful. 
In our case, Text2Topic is a better choice because of: 
1) less dependency on non-open-source models, so that the model iteration and rate limits are under-control; 
2) avoiding tedious prompt tuning procedures; 
3) a lower cost and it's more eco-friendly: the Text2Topic model has less than 200M parameters (considering we cache the topics embedding during inference). If the GPT-3.5 has 175B parameters, it would be more than 800 times bigger. As described in Section \ref{sec:deploy_batch}, Text2Topic can reach 8000 text/min throughput, with a \$7.5 cost predicting on 1M text, while GPT-3.5 would cost \$6250 (assuming the prompt and text have 3.5K tokens, the output (CoT and topic list result) has 500 tokens) \footnote{OpenAI price on gpt-3.5-turbo-0301 is \$0.0015 / 1K input tokens + \$0.0020 / 1K output tokens, when writing this paper. }. 
4) better scalability for larger number of topics and less worrying about prompting length exceeding certain limitation. Though we can split topics into multiple groups/prompts and do multiple calls on GPT-3.5, it means a higher cost and the group split setting is difficult to optimize.



\subsection{Model Interpretation} \label{sec:lime}

We use Local Interpretable Model-agnostic Explanations (LIME) \cite{lime} for model interpretation. LIME generates local explanations by perturbing individual text instances, 
approximating the model behavior by using a surrogate model that highlights the importance of words in the original text. 
LIME is effective in the error-analysis flow, and help topic description refinements at an early stage (see example in Figure \ref{fig:explanation-exp} in the Appendix).

\section{Real-World Applications} \label{sec:use_case}


This section describes 3 main real-world use cases which employ Text2Topic. 
Besides, the Text2Topic training system is also re-used effectively for training other data sources like search queries. 

It is important to note that use cases may require varying threshold settings. We use the F-beta score \cite{fbeta} to determine the optimal threshold setting on the topic's probability score for a given use case. For recall-oriented scenarios, where minimizing false negatives is critical, we set beta > 1. Conversely, for precision-oriented cases, where reducing false positives is a priority, we set beta < 1. 
For each topic, to find the best threshold, we systematically vary its value by computing the threshold that yields the highest F-beta score, using the chosen beta value. 



\textbf{Property Recommendation}: 
Reviews contain rich information that encapsulates the 
users' preferences towards different properties. Text2Topic turns them into structured features, which enhance the in-house property recommendation models' performance. 
With classification scores on reviews, we perform property-level score aggregations, to extract a variety of insightful attributes such as a property's relevancy for different themes (e.g., beach, spa/wellness). These attributes are integrated into the recommendation models to increase relevant inventory (e.g., number of beach properties is increased by 4\% by leveraging Text2Topic); and to create novel and nuanced categories of recommendations (such as castle-type hotels, romantic getaways, etc.).
Furthermore, the model provides a natural mechanism to serve explainable recommendations by linking them to relevant reviews.

\textbf{Detect Property Type}: 
With Text2Topic predictions on reviews, we are able to detect hidden properties categorizations, by analyzing relevant topic (guest house, farm stay, resort, chalet etc.) frequencies. 
For example, an Apartment property that is described as a 
Guest house, could be surfaced to users that are searching for a guest house. 
We detect over a million extra properties supply (774K more apartments, 25k more villas and 60k more cabins/chalets). 



\textbf{Fintech}:
Text2Topic training pipeline enables us to train a new model on Fintech data and topics, such as payments and questions about invoices and commissions. 
The model auto-classifies incoming messages from customers and correctly re-routes them to the right self-service solution, which increases the auto-reply success rate by $9\%$ and reduces manual handling time. 



\section{Deployment} \label{sec:deploy}

\subsection{The Deployment Platform}
The model is deployed and monitored on a stream processing platform 
based on Apache Flink \citep{apache}.  
It consumes real-time events from Kafka \citep{kafka} topics to generate model-based predictions. It automatically scales up the number of model endpoints to better handle peak times and allows leveraging Apache Flink’s asynchronous I/O operator to perform concurrent asynchronous HTTP calls to the model endpoint. 
The architectural design allows the platform to be also used for backfilling (scoring historical data with newly deployed model), by simulating events of historical data and pushing them to Kafka.  
The platform is designed to achieve high prediction throughput while keeping a low latency. 


\subsection{Model Serving and Batch Invocations} \label{sec:deploy_batch}
To maximize the hardware utilization (NVIDIA Tesla V100 GPUs for production), we combine batch model invocations with Flink's native asynchronous I/O support. For batch model invocations, we leverage Flink’s windowing mechanism to implement the grouping of events that will be sent to the model together in a single API call.
Events are accumulated to windows as soon as they become available for consumption from the source Kafka topic.
The window is closed after a predefined time period (e.g. 3 seconds), or whenever the number of accumulated events reaches the desired batch size. 

Aiming to minimize the cost per prediction, we start with grid searching the optimal batch size by performing stress tests using a single model endpoint. For each batch size we randomly sample batches of texts from the corpus, and then iterate over the batches sequentially and invoke the model. 
We observe that while increasing the batch size, the throughput (number of texts predictions per minute) first increases, and then starts decreasing as the number of available GPU cores exhausted. An optimal and memory-explosion safe batch size is 300. 
Then
we run experiments to compare batch invocations against asynchronous I/O invocations. As Table \ref{tab:throughput} in the Appendix shows, using synchronous API calls with a batch size of 300 yields a cost of \$7.5 for 1M predictions, while using 50 concurrent asynchronous I/O API calls without batching yields a cost of \$15. So the former is selected. In addition, for backfilling, texts with similar lengths are grouped together and we apply dynamic padding to the longest element in each batch, which reduces computational overhead. 


\section{Conclusion}
In this paper, we present Text2Topic, a flexible multi-label text classification system that is deployed at Booking.com, with high performance and supports multiple applications. We summarize lessons learnt from the end-to-end production journey, including practical annotation approaches, modeling choices, and production decisions, which are valuable references for the industry domain. We also compare the performance with LLM like GPT-3.5, and Text2Topic is a more feasible choice from multiple aspects. For future work, we can explore if parameter efficient fine-tuning techniques (e.g., LoRA \cite{lora}, p-tuning \cite{ptuning2}) on open-source LLMs could bring better performance, and how to better balance the model specialization and generalization power for zero-shot. 






\section*{Acknowledgments}
This work is supported by Booking.com. We would like to thank Satendra Kumar, Selena Wang, Michael Alo, and Guy Nadav for the paper review. We would also like to thank Ilya Gusev on contributing some GPT-3.5 prompting ideas. 

\bibliography{custom}
\bibliographystyle{acl_natbib}

\clearpage
\appendix

\section{Appendix}

\begin{table}[!h]
\resizebox{1\columnwidth}{!}{
\begin{tabular}{lccc}
\hline
\textbf{method}                & \textbf{\begin{tabular}[c]{@{}c@{}}macro \\ mAP\end{tabular}} & \textbf{\begin{tabular}[c]{@{}c@{}}weighted \\ mAP\end{tabular}} & \textbf{\begin{tabular}[c]{@{}c@{}}macro \\ F1 score\end{tabular}} \\ \hline
MUSE                           & 54.4                                                          & 58.3                                                             & 57.6                                                                 \\ \hline
Bi-encoder (cosine)            & 47.4                                                          & 62.4                                                             & 51.6                                                                 \\ \hline
Bi-encoder (concat, sub, mult) & \textbf{58.1}                                                 & \textbf{68.5}                                                    & \textbf{60.0}                                                        \\ \hline
\end{tabular}
}
\caption{Zero-shot overall test-set performance on popular topics which have more than 50 positive annotations each in the test set. We search the best F1 across all thresholds per topic, and then get macro averaged F1 across topics. }
\label{tab:zero-shot-top-50}
\end{table}

\begin{table}[!h]
\resizebox{1\columnwidth}{!}{
\begin{tabular}{cccc}
\hline
\textbf{\begin{tabular}[c]{@{}c@{}}Batch \\ Size\end{tabular}} & \textbf{\begin{tabular}[c]{@{}c@{}}\#Concurrent \\ Calls \\ (Async I/O)\end{tabular}} & \textbf{\begin{tabular}[c]{@{}c@{}}Throughput\\ (\#texts per \\ minute)\end{tabular}} & \textbf{\begin{tabular}[c]{@{}c@{}}\$USD/1M \\ Predictions\end{tabular}} \\ \hline
1                                                              & 50                                                                                           & 4,000                                                                                    & \$15                                                                            \\
300                                                            & 3                                                                                            & 7,200                                                                                    & \$8.3                                                                           \\
300                                                            & 1                                                                                            & 8,000                                                                                    & \textbf{\$7.5}                                                                  \\ \hline
\end{tabular}
}
\caption{Comparing batch model invocations and Async I/O approach. Maximizing the batch size doubles the model invocation throughput and reduces the prediction costs by half. Combining too many asynchronous calls with a large batch size exhausted the GPU resources, which resulted in a reduced throughput compared to pure batch invocations. } 
\label{tab:throughput}
\end{table}

\begin{figure}[!h]
    \centering
    \includegraphics[width=1\columnwidth]{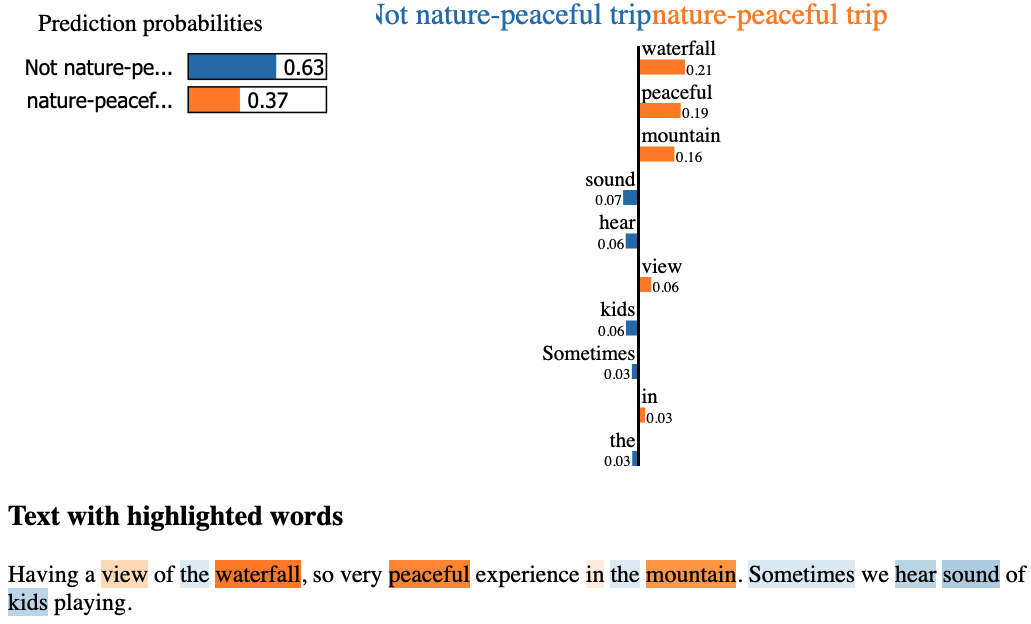}
        \caption{LIME explanation for a specific text-topic pair. We can see word-level importance in detecting ``nature-peaceful trip'' topic. Orange color indicates positive influence (words like: view, waterfall, peaceful, mountain) and Blue color indicates negative influence (words like: kids, sound). LIME explanation helps human to refine topic descriptions, for example by removing unclear wording, or adding more precise and concise wording, at the early stage of this project.}
    \label{fig:explanation-exp}  
\end{figure}

\begin{figure}[!h]
    \centering
    \includegraphics[width=1\columnwidth]{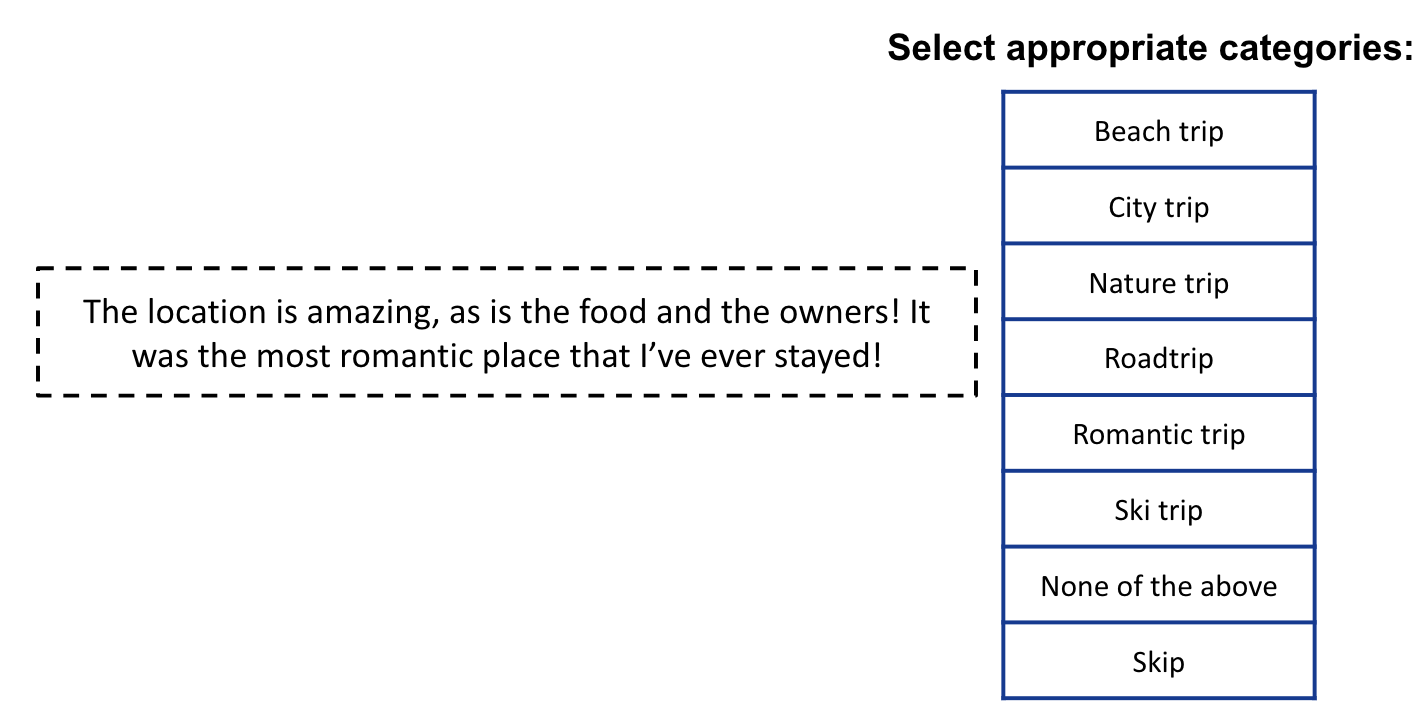}
        \caption{Example of the topics group assigned to a text in an annotation task. The annotator can select multiple topics.}
    \label{fig:annotation_example}  
\end{figure}

\begin{figure*}
    \centering
    \includegraphics[width=0.9\textwidth]{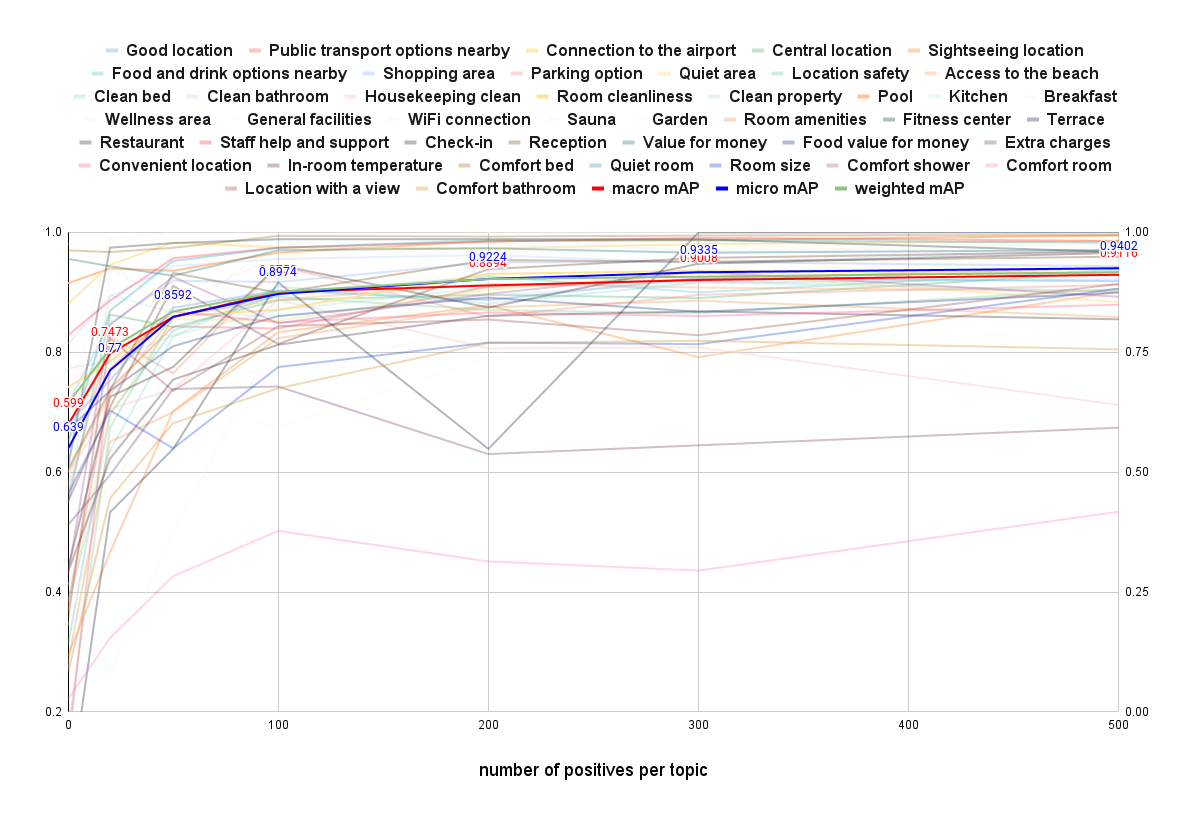}
        \caption{Performance tracking when sampling different number of positive annotations per topic for model training. This provides a general guide: for most topics, 200 number positive annotations is enough. However, for training hundreds of topics in one model, we might need more positive annotations per topic, so we also consider it when deciding on the annotation volume. }
    \label{fig:num_anno}
\end{figure*}

\begin{figure*}
    \centering
    \includegraphics[width=0.8\textwidth]{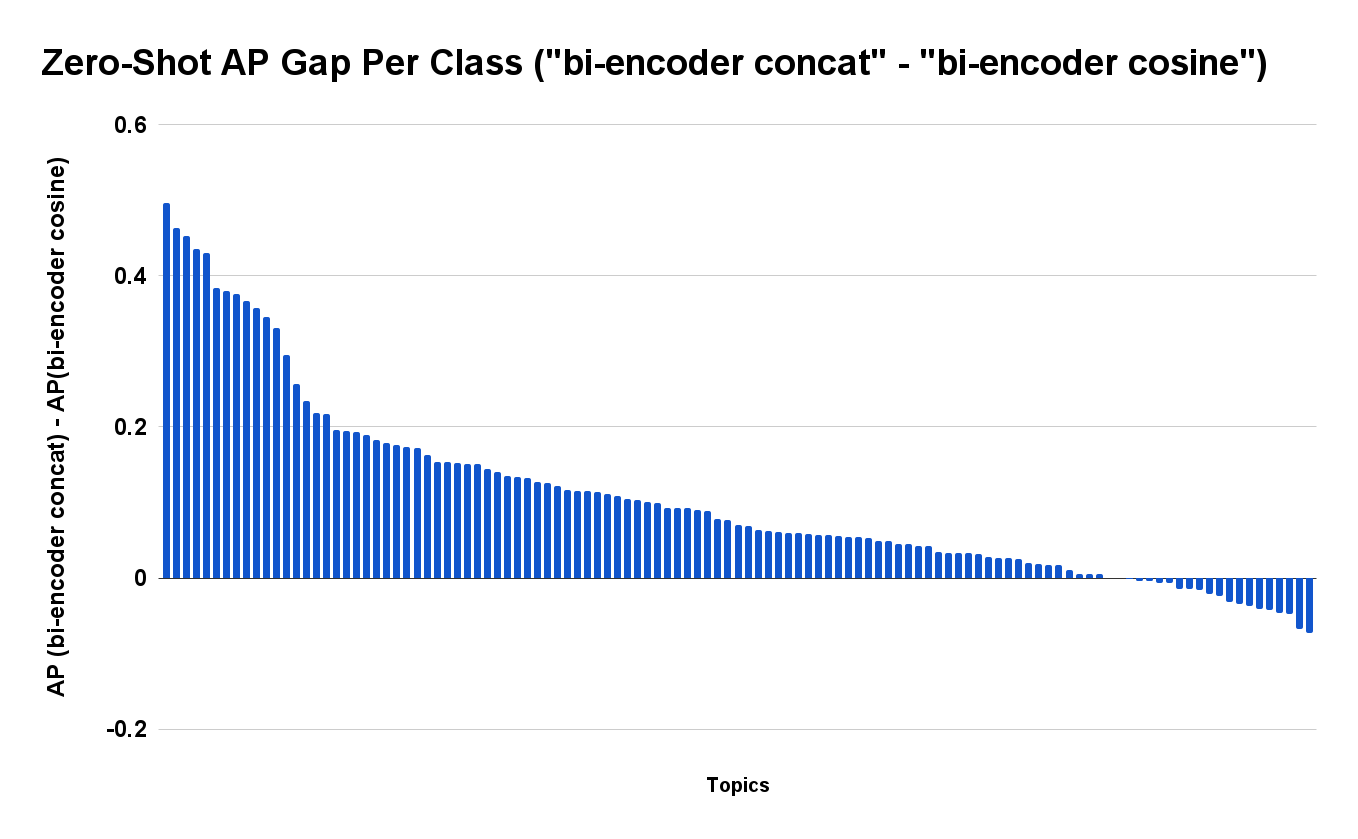}
        \caption{Zero-shot Average Precision score gap on each topic, on the test set. 
        The x-axis represents the topics, ordered by the score gap. 
        The y-axis shows the gap between the AP score of ``bi-encoder concat'' model and ``bi-encoder cosine'' model. 
        This plot includes the popular topics which have more than 50 positive annotations each in the test set. 
        We can see the cosine one is almost always worse than the concat one. 
        }
    \label{fig:zero_shot}
\end{figure*}

\begin{figure*}
    \centering
    \includegraphics[width=0.8\textwidth]{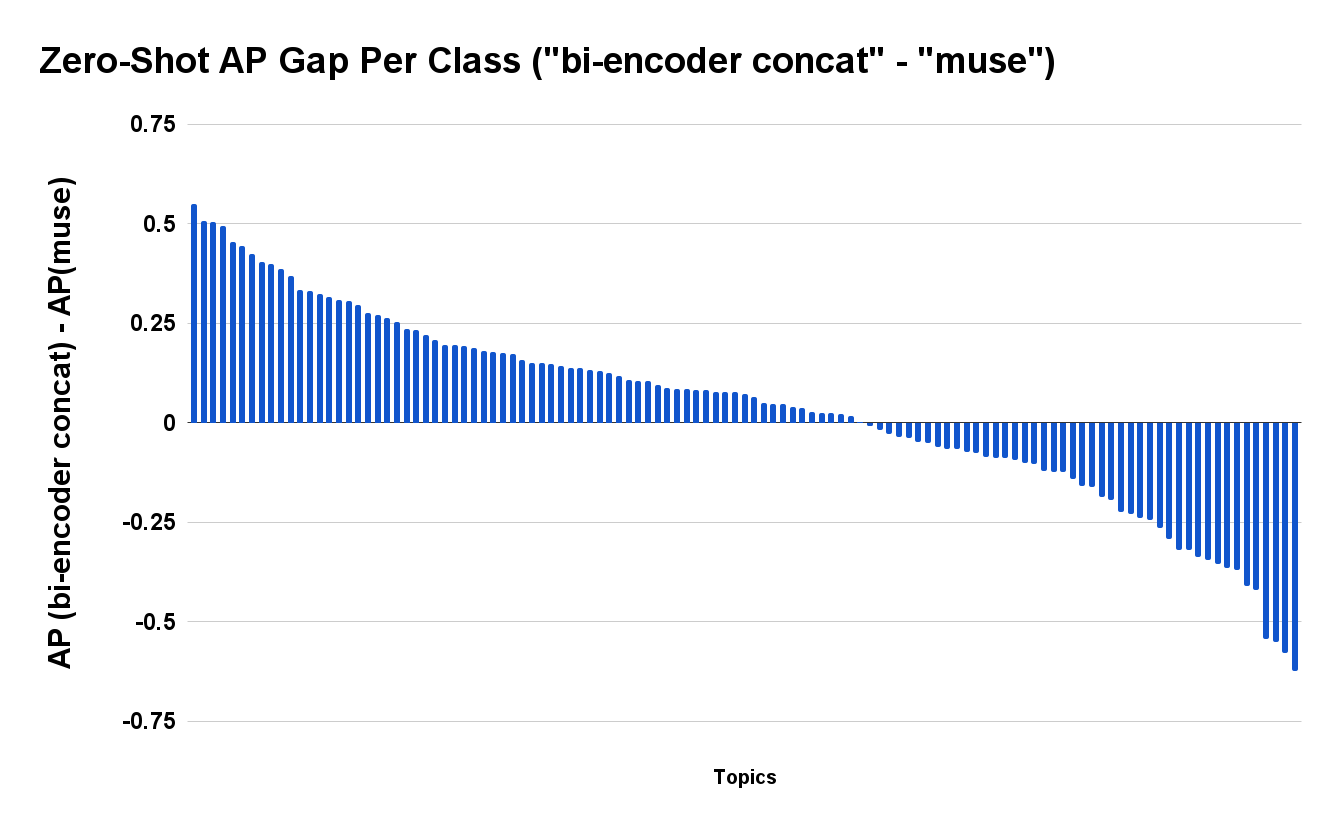}
        \caption{Zero-shot Average Precision score gap on each topic, on the test set. 
         The x-axis represents the topics, ordered by the score gap.
        The y-axis shows the gap between the AP score of ``bi-encoder concat'' model and MUSE. 
        This plot includes the popular topics which have more than 50 positive annotations each in the test set. 
        We can see that when inspecting topic-level performance, the Text2Topic bi-encoder concat has stronger zero-shot ability than MUSE. MUSE is better at some topics, which we find are mainly facility specific topics (BBQ, towel, vending machine, stairs etc.). 
        }
    \label{fig:zero_shot_rank_on_muse}
\end{figure*}





\end{document}